\documentclass[11pt,a4paper]{article}
\usepackage[hyperref]{acl2020}
\usepackage{times}
\usepackage{latexsym}

\usepackage{latexsym}
\usepackage{amsmath}
\usepackage{multirow}
\usepackage{graphicx}  
\usepackage{tabularx}
\usepackage{ragged2e}
\usepackage{booktabs}
\usepackage{enumitem}
\setlist[enumerate]{wide=0pt}

\usepackage{microtype}

\aclfinalcopy 
\def\aclpaperid{2518} 

\title{Towards Robustifying NLI Models Against Lexical Dataset Biases}

\author{Xiang Zhou $\;\;\;\;$ Mohit Bansal \\
  UNC Chapel Hill \\
  \texttt{\{xzh, mbansal\}@cs.unc.edu} \\
}

\usepackage{xcolor}

\date{}
\begin{document}

\maketitle

\begin{abstract}
While deep learning models are making fast progress on the task of Natural Language Inference, recent studies have also shown that these models achieve high accuracy by exploiting several dataset biases, and without deep understanding of the language semantics. Using contradiction-word bias and word-overlapping bias as our two bias examples, this paper explores both data-level and model-level debiasing methods to robustify models against lexical dataset biases. First, we debias the dataset through data augmentation and enhancement, but show that the model bias cannot be fully removed via this method. Next, we also compare two ways of directly debiasing the model without knowing what the dataset biases are in advance. The first approach aims to remove the label bias at the embedding level. The second approach employs a bag-of-words sub-model to capture the features that are likely to exploit the bias and prevents the original model from learning these biased features by forcing orthogonality between these two sub-models. We performed evaluations on new balanced datasets extracted from the original MNLI dataset as well as the NLI stress tests, and show that the orthogonality approach is better at debiasing the model while maintaining competitive overall accuracy.\footnote{Our code and data are available at: \url{https://github.com/owenzx/LexicalDebias-ACL2020}}
\end{abstract}
\vspace{-3pt}

\begin{table*}[t!]
\small
\begin{tabularx}{\textwidth}{|l|l|X|X|}
\hline
 &  &\bf Contradiction-Word Bias &\bf Word-Overlapping Bias \\ \hline
\multirow{2}{*}[-1em]{\rotatebox{90}{\bf MNLI}} & Prem. & A recorded menu will provide information on how to obtain these lists. & This is especially true on Menocra, where cold winter winds limit the season’s length. \\ \cline{2-4} 
 & Hypo. & Recorded menus do\textbf{ not} provide any information at this time. & On Menocra, where cold winter winds limit the season’s length, this is especially true. \\ \hline
\multirow{2}{*}[-1em]{\rotatebox{90}{\bf Stress}} & Prem. & Understanding is the key.  & This is especially true on Menocra, where cold winter winds limit the season’s length. \\ \cline{2-4} 
 & Hypo. & Understanding is the most important \textbf{and false is not true}. & On Menocra, where cold winter winds limit the season’s length, this is especially true \textbf{and true is true}. \\ \hline
\end{tabularx}
\vspace{-6pt}
\caption{The example samples for the CWB and WOB in the original dataset and the test samples in NLI stress tests \cite{naik2018stress} designed to reveal these biases (the stress test samples aim to fool the model to predict contradiction by adding negation word and to not predict entailment by reducing word overlapping). }
\label{tab:example}
\vspace{-9pt}
\end{table*}

\section{Introduction}
\label{sec:intro}
In this work, we focus on investigating and reducing biases in the task of Natural Language Inference (NLI), where the target of the model is to classify the relations between a pair of sentences into three categories: entailment, neutral and contradiction. With the release of large-scale standard datasets \cite{bowman2015large,williams2018broad}, significant success has been made on this task, and recent state-of-the-art neural models have already reached competitive performance even compared to humans. 
However, a number of papers \cite{gururangan2018annotation,poliak2018hypothesis,nie2018analyzing,naik2018stress} have shown that despite the high accuracy on these datasets, these models are far from mastering the required nature of natural language inference. Instead of deeply understanding the sentences in the correct semantic way, these models tend to exploit shortcuts or annotation artifacts in the dataset and actually overfit to these datasets to predict the label using simple patterns. However, most shortcuts are only valid within the datasets and fail to hold for general natural language. Hence, these models fail to generalize to other datasets for the same task~\cite{DBLP:journals/corr/abs-1810-09774}, perform badly on challenge analysis datasets \cite{glockner2018breaking, mccoy2019right, wang2019if}, and are fooled by adversarial attacks \cite{naik2018stress}.

One major cause of this problem is the existence of dataset biases. Since most NLP datasets are often collected and processed by crowdworkers, bias can be added to the data at every step of data collection. For example, when writing contradiction pairs, workers are likely to use negation words such as `not', and when creating entailment pairs, workers are likely to keep most of the words in the premise sentence. This results in `annotation artifacts' in the dataset~\cite{gururangan2018annotation}. In reality, almost every dataset contains countless such diverse biases. In our paper, we focus on the Multi-Genre Natural Language Inference (MNLI) dataset \cite{williams2018broad} in English, and on two specific kinds of dataset bias:

\noindent\textbf{Contradiction Word Bias (CWB)}: If the hypothesis sentence contains some specific words (such as negation words) that are always used by the crowd-workers to generate contradiction pairs, then the sentence pair is very likely to be contradiction.

\noindent\textbf{Word Overlapping Bias (WOB)}: If the premise sentence and the hypothesis sentence have a high word-overlap, then the sentence pair is very likely to be entailment.

These two types of biases are selected as the focus of our experiments because: (1) there exist a significant number of samples in the dataset where they are a major problem; (2) they are conceptually easy to understand and relatively easier to evaluate. In our experiments, we not only used current existing evaluation datasets from~\citet{naik2018stress}, but also extracted balanced evaluation datasets from the original data to evaluate these two biases. Although we only focus on these two kinds of dataset biases throughout our experiments, our methods are not specifically designed for these two biases and should be able to reduce other similar lexical biases simultaneously.

Using these two example lexical biases, our paper discusses the following three questions:
\begin{itemize} 
\vspace{-5pt}
\itemsep-0.5em 
    \item[Q1.] Is lexical bias a problem that can be solved by only balancing the dataset?
    \item[Q2.] Can the lexical bias problem be solved using existing ideas from the gender bias problem?
    \item[Q3.] What are some promising new modeling directions towards reducing lexical biases? 
\end{itemize}
\vspace{-3pt}

As responses to these three questions, we conduct three lines of experiments. Firstly, we expand the discussion of Q1 by studying whether and how the bias can be reduced by debiasing the dataset. For this, we add new training data which does not follow the bias pattern. This new data can come from two sources, either from the original training set or via manually generated synthetic data. We show that both methods can slightly reduce the model's bias. However, even after adding a large amount of additional data, the model still cannot be completely bias-free. Another critical problem with these data augmentation/enhancement based debiasing methods is that we need to know the specific behaviour of the biases before making some related changes to the dataset. However, in reality, models are always faced with new training datasets containing unknown and inseparable biases. Hence, the answer to Q1 is mostly negative for simple data-level approaches and we also need to focus on designing direct model-debiasing methods.

Therefore, we turn our focus to directly debiasing the model (Q2 and Q3). 
The first method is to debias the model at the lower level, i.e., by directly debiasing the embeddings so that they do not show strong biases toward any specific label. This is one of the most prevalent methods for reducing gender biases, so through the examination of this idea, we aim to compare lexical bias problems to gender bias problems and highlight its uniqueness (hence answering Q2). Finally, we debias the model at the higher level, i.e., by designing another bag-of-words (BoW) sub-model to capture the biased representation, and then preventing the primary model from using the highly-biased lexical features by forcing orthogonality between the main model and the BoW model (via HEX projection \cite{wang2018learning}). In our experiments, we show that debiasing the prediction part of the model at higher levels using BoW-orthogonality is more effective towards reducing lexical biases than debiasing the model's low-level components (embeddings). This approach can significantly robustify the model while maintaining its overall performance, hence providing a response to Q3.  We also present qualitative visualizations using LIME-analysis for the important features before and after applying the BoW-orthogonality projection.

\section{Related Work}
\label{sec:related_work}

\vspace{2pt}
\noindent\textbf{Problems with NLI Models and Datasets.} Despite the seemingly impressive improvements in NLI tasks, recently a number of papers revealed different problems with these models. \citet{gururangan2018annotation} showed that annotation artifacts in the datasets are exploited by neural models to get high accuracy without understanding the sentence. \citet{poliak2018hypothesis} showed a similar phenomenon by showing models getting good performance but only taking one sentence as the input. \citet{nie2018analyzing} showed that NLI models achieved high accuracy by word/phrase level matching instead of learning the compositionality.
\citet{naik2018stress} constructed bias-revealing datasets by modifying the development set of MNLI. In our evaluation, besides using the datasets from \citet{naik2018stress}, we also extract new datasets from the original MNLI dataset to maintain the consistency of input text distribution.

\vspace{2pt}
\noindent\textbf{Adversarial Removal Methods.} Adversarial removal techniques are used to control the content of representations. They were first used to do unsupervised domain adaptation in \citet{ganin2015unsupervised}. \citet{xie2017controllable} later generalized this approach to control specific information learned by the representation. \citet{li2018towards} used a similar approach to learn privacy-preserving representations. However, \citet{elazar2018adversarial} showed that such adversarial approach fails to completely remove demographic information. \citet{minervini2018adversarially} generate adversarial examples and regularize models based on first-order logic rules. \citet{belinkov-etal-2019-adversarial, belinkov2019don} showed that adversarial removal methods can be effective for the hypothesis-only NLI bias. Our focus is on two different lexical biases and our results are complementary to theirs.\footnote{We have tried a similar approach via gradient reversal w.r.t. BoW sub-model in preliminary experiments and observed less effectiveness (than HEX-projection), which hints that different types of biases can lead to different behaviors.} Recently, \citet{wang2018learning} proposed HEX projection to force the orthogonality between the target model and a superficial model to improve domain generalization for image classification tasks. Here, to make the model less lexically biased, we apply the HEX projection with specially-designed NLP model architectures to regularize the representation in our models. Even more recently, \citet{clark2019don} and \citet{he2019unlearn} propose to robustify the task model with the help of an additional simple model, using ensembling to encourage cooperation of the two models. On the other hand, our main motivation to compare the advantages/limitations of dataset vs. embedding vs. classifier debiasing methods (against two different types of problematic lexical biases in NLI), and also our classifier debiasing method forces the task model to capture orthogonal information via HEX projection.

\vspace{2pt}
\noindent\textbf{Removing Gender Bias in NLP Models.} There is also a line of work in NLP on analyzing and reducing gender bias in NLP models. \citet{NIPS2016_6228,caliskan2017semantics,zhao2018gender} studied the bias problem in word embeddings. \citet{zhao2017men} reduced gender bias in visual recognition using corpus-level constraints. \citet{zhao2018learning} discussed the gender bias problem in co-reference resolution. These problems are related to our work, but lexical biases are more complex. Multiple inseparable lexical dataset biases can influence one single example and the same word can have different lexical biases in different contexts. Later in our experiments, we show that these two problems behave differently and we present the need for different solutions.

\section{Data-Level Debiasing}
\label{sec:debais_dataset}
Models naturally learn the biases from the dataset they are trained on. Therefore, as we mentioned in Q1 in Sec.~\ref{sec:intro}, one may first wonder if lexical bias can be completely removed by fixing the source of the bias, i.e., datasets.
While collecting large-scale datasets~\cite{bowman2015large,williams2018broad} already takes a lot of time and effort, collecting bias-free datasets is even more time-consuming and hard to control. Therefore, here we focus on getting additional data from currently-available resources. We conducted experiments using two resources of data. The first one is to do `data enhancement' by repeating samples in the original training data. The second source is `data augmentation' by manually creating synthetic data. We follow the construction of existing synthetic bias-revealing datasets to create new samples for the training set so that these targeted biases can be reduced. 

\noindent\textbf{Data Enhancement by Repeating Training Data.} For most kinds of biases, there still exists a small portion of samples that don't follow the bias. Therefore, we reduce biases in datasets by repeating this portion of samples. For CWB, we select non-contradiction samples containing contradiction words (details see Sec. \ref{sec:datasets}) in the hypothesis sentence but not in the premise sentence. For the WOB, we select non-entailment samples with highest word overlapping (measured by the Jaccard distance \cite{hamers1989similarity} of words).  Next, since the number of these unbiased samples may not be large enough, we repeatedly add those selected samples to make the training set more balanced. The results from adding 500 new samples to 50,000 new samples are shown in Sec.~\ref{sec:result_data}.

\noindent\textbf{Data Augmentation by Adding Synthetic Data.} Researchers have been using synthetic rules to generate harder or perturbed samples to fool the model. Here, besides using these datasets only as the evaluation set, we also add these samples back to the training set, similar to the concept of adversarial training \cite{jia2017adversarial,wang2018adversarial,niu2018adversarial} where the adversarial examples are added back to the training set so that the resulting model will be more robust to similar adversarial attacks. In our experiments, we follow \citet{naik2018stress} to append meaningless sentences at the end of the hypothesis sentence like in Table \ref{tab:example} to create additional new samples. The detailed construction of these samples can be seen in Appendix. By learning from these augmented datasets, the model should also be more robust to certain types of perturbations/biases of the data. 

In Sec.~\ref{sec:result_data}, our experiments showed that while this approach can lead to less biased models, it cannot make the model completely bias-free. 
Another disadvantage of these data enhancement/augmentation approaches is that we need to know all the specific kinds of biases in advance. For instance, in order to reduce the CWB for `not', one needs carefully balance the samples containing `not' in the training set. However, lots of other words will exhibit similar biases (e.g., the model tends to predict neutral when it sees `also') and it is impractical to identify and debias the dataset w.r.t. every type of bias. Therefore, besides fixing the dataset, we should also focus on directly debiasing models against lexical biases.

\section{Model-Level Debiasing}
\label{sec:debais_model}
Model-level debiasing methods have the advantage that there is no need to know the specific bias type in advance.
Here we propose two different methods. The first method focuses on debiasing the content of word/sentence embeddings, where we aim to remove strong bias in the embeddings towards any of the labels  so that there will be fewer shortcuts for models to exploit. The second method builds a separate shallow bag-of-words (BoW) sub-model and projects the primary model's representation onto the subspace \emph{orthogonal} to this BoW sub-model via the HEX projection algorithm~\cite{wang2018learning}. Our proposed methods can be applied to a wide range of  baseline model architectures. In addition, none of our methods is bias-type specific, so the results on CWB and WOB should generalize to other similar lexical biases. 

\subsection{Baselines}
We use sentence-embedding based models as our baseline since they are more controllable, and because the interaction of sentences only appears at the top classifier, which makes it easier to compare the different effects of different regularization.\footnote{Another popular choice of NLI model architecture is the cross-attention based models \cite{chen2017enhanced, devlin2018bert}. In our current work, we choose to only apply our BoW Sub-Model approach on sentence-embedding based models since our approach directly regularizes the representation vector learned by the main model, and hence it is most suitable for models with a single vector containing rich information. On the other hand, cross-attention based models do most of the inference through cross-attention and do not learn such a single vector, making it hard to regularize the model effectively in a similar way. Investigation of similar HEX regularization methods for cross-attention models is future work.}
Our baseline structures can be divided into three stages. The first stage is to embed the words into word embeddings. The second stage is to get the representations for each sentence. We use three layers of BiLSTM to get the representation. We also added residual and skip-connections as \citet{nie2018analyzing}, and find that it leads to better performance.
For the final stage, our baseline follows \citet{mou2016natural,conneau2017supervised} to concatenate these two sentence embeddings, their difference, and their element-wise product as follows:
\begin{equation} 	\label{equ:concat}
    m = [h_1;h_2;h_1-h_2;h_1\odot h_2]
\end{equation}
The resulting vector is passed through another multi-layer perceptron (MLP) to get the final classification result.\footnote{Our baseline models achieve close to the best sentence embedding based/cross-attention based models reported on the NLI stress tests \cite{naik2018stress} and are hence good starting points for this bias/debias analysis.} 

Next, we will describe two different methods to directly debias the model.

\begin{figure}[t]
	\includegraphics[width=0.94\linewidth]{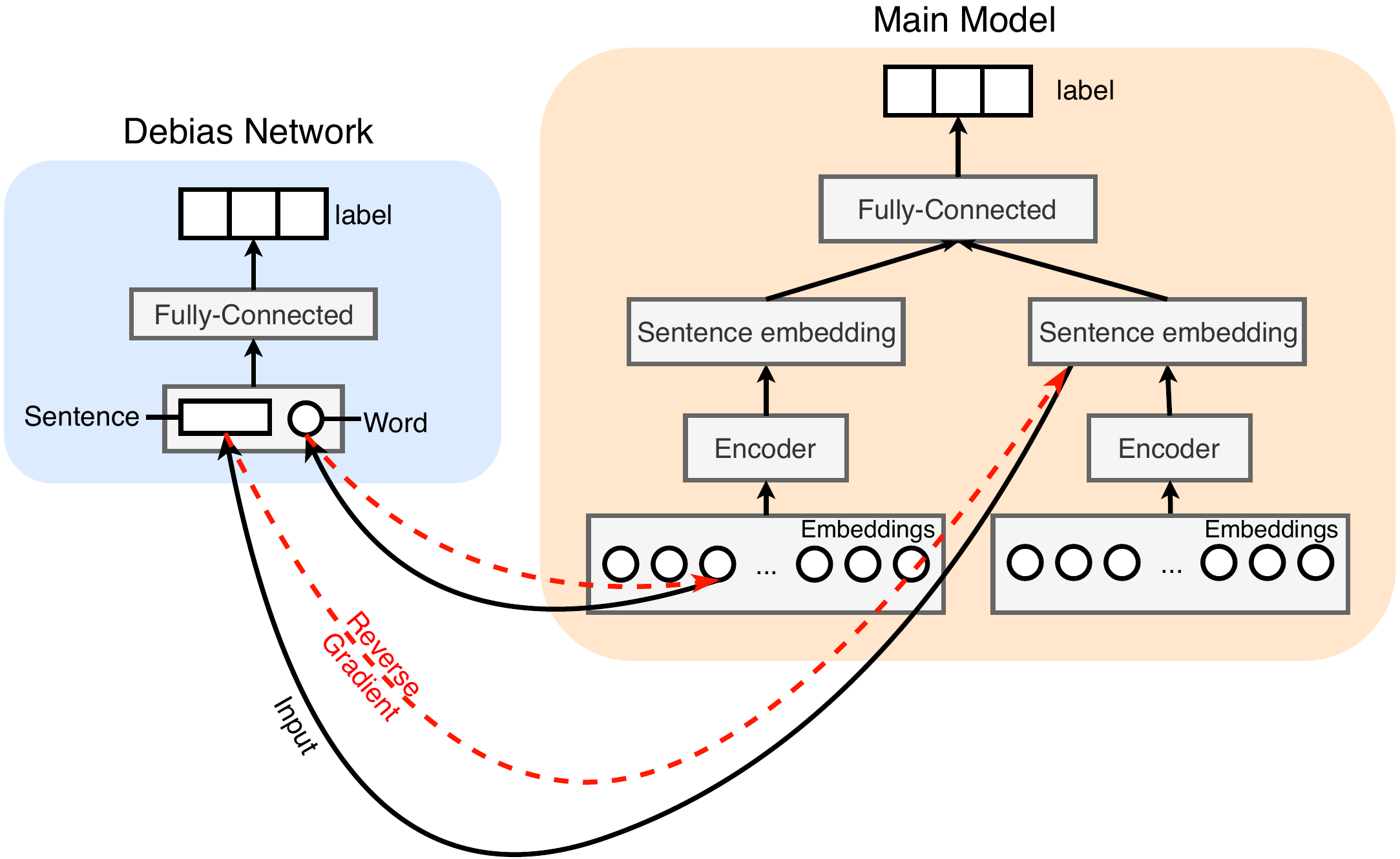}
	\vspace{-8pt}
	\caption{The overall architecture for reducing bias using an embedding debiasing network. The red dashed line denotes gradient reversal. \label{fig:emb_debias}}
	\vspace{-10pt}
\end{figure}

\subsection{Debiasing Embeddings}
Word embeddings are an important component in all neural NLP models. They contain the most basic semantics of words. Recent studies have shown that removing gender bias from word embeddings can lead to less biased models~\cite{zhao2018gender}. In our work, as we discussed in Q2 in Sec.~\ref{sec:intro}, we explore whether similar ideas can be applied to reducing lexical dataset biases.

For a large number of lexical dataset biases (e.g., CWB), the model tends to predict the label based only on the existence of certain words. Hence, one natural conjecture is that there is a strong bias towards some labels in the word embeddings. Since the label bias is not an attribute of the word, but it is brought in by the model above, hence in order to remove such label bias from the embeddings at training time, we differ from  \citet{zhao2018gender} to use the gradient-reversal trick \cite{ganin2015unsupervised,xie2017controllable}.

The architecture of this approach is illustrated in Figure \ref{fig:emb_debias}. We denote the embeddings of the two input sequences for our model as $\mathbf{w^{(a)}}=\{\mathbf{w}^{(a)}_1,\mathbf{w}^{(a)}_2,\dots,\mathbf{w}^{(a)}_{l_a}\}$ and $\mathbf{w^{(b)}}=\{\mathbf{w}^{(b)}_1,\mathbf{w}^{(b)}_2,\dots,\mathbf{w}^{(b)}_{l_b}\}$ respectively, where $a$ denotes the premise sentence while $b$ denotes the hypothesis sentence.
In order to apply the reverse gradient trick \cite{ganin2015unsupervised} to the embeddings, we add a small embedding-debias network (the left blue box in Figure \ref{fig:emb_debias}) for each of the embedding $w_i$ in our model. The embedding-debias network is a simple MLP.
Since the other parts of the sentence context may also contribute to the bias, the debiasing network takes both $w^{(a)}_i$ and the sentence embedding of b (and vice versa for debiasing $w^{(b)}$) as the input and  predicts the label $y$.
Therefore, the total loss of this method is:
\begin{equation*}
\vspace{-3pt}
    L(\mathbf{\theta}_c, \mathbf{\theta}_e, \mathbf{\theta}_{ed})=L_c(\mathbf{\theta}_c, \mathbf{\theta}_e)-\frac{\lambda}{l_a+l_b}L_{ed}(\mathbf{\theta}_e, \mathbf{\theta}_{ed})
\end{equation*}
Here, $\lambda$ is the multitask coefficient. $l_a$ and $l_b$ are the lengths of two input sentences. $L_c$ is the standard classification loss using the main model and $L_{ed}$ is the sum of all the classification loss using the debias network. $\mathbf{\theta}_e$ are parameters of the embeddings and sentence encoder of the main model, $\mathbf{\theta}_c$ are parameters of the top classifier of the main model, and $\mathbf{\theta}_{ed}$ are parameters of the embedding-debias network. In order to find the optimal parameters, we follow \citet{ganin2015unsupervised} to reverse the gradient for $\mathbf{\theta}_{e}$ w.r.t. $L_{ed}$.

 Besides this approach, we also tried two variants by changing the input of the debias network. The first one is \textbf{emb\_basic}, where we only take the single embedding $w_i$ as the input. The second one only takes one sentence embedding as the input and is called \textbf{ind\_sent}.
 The results of our embedding-debias methods are shown in Sec. \ref{sec:result_model}.

\subsection{Bag-of-Words Sub-Model Orthogonality}
While debiasing the embeddings can robustify the models against certain biases, it may not be effective for all the lexical biases. Some lexical bias may exist at the deeper compositionality level (e.g., WOB), while debiasing the embeddings can regularize only the most basic semantics units instead of how these semantics units are composed by the model. In addition, removing the label biases may also hurt the useful semantics contained in the embeddings, leading to significant performance drops. A better approach is to leave the embedding intact, but try to regularize how the classifier uses these features. We observe that models exploiting dataset biases in the training set (e.g., CWB and WOB) tend to use very simple and superficial features to make the prediction. These models tend to ignore the order of the words, fail to learn compositionality, and do not have a deep semantic understanding of the sentences.
Therefore, we aim to robustify the model by letting it use fewer simple and superficial features. With this motivation, we train a bag-of-words (BoW) model that only captures superficial patterns of the words without any word order/compositionality information. Then we use HEX projection \cite{wang2018learning} to project the representation of the original primary model to the orthogonal space of the representation of the BoW model.

\begin{figure}[t]
	\includegraphics[width=0.94\linewidth]{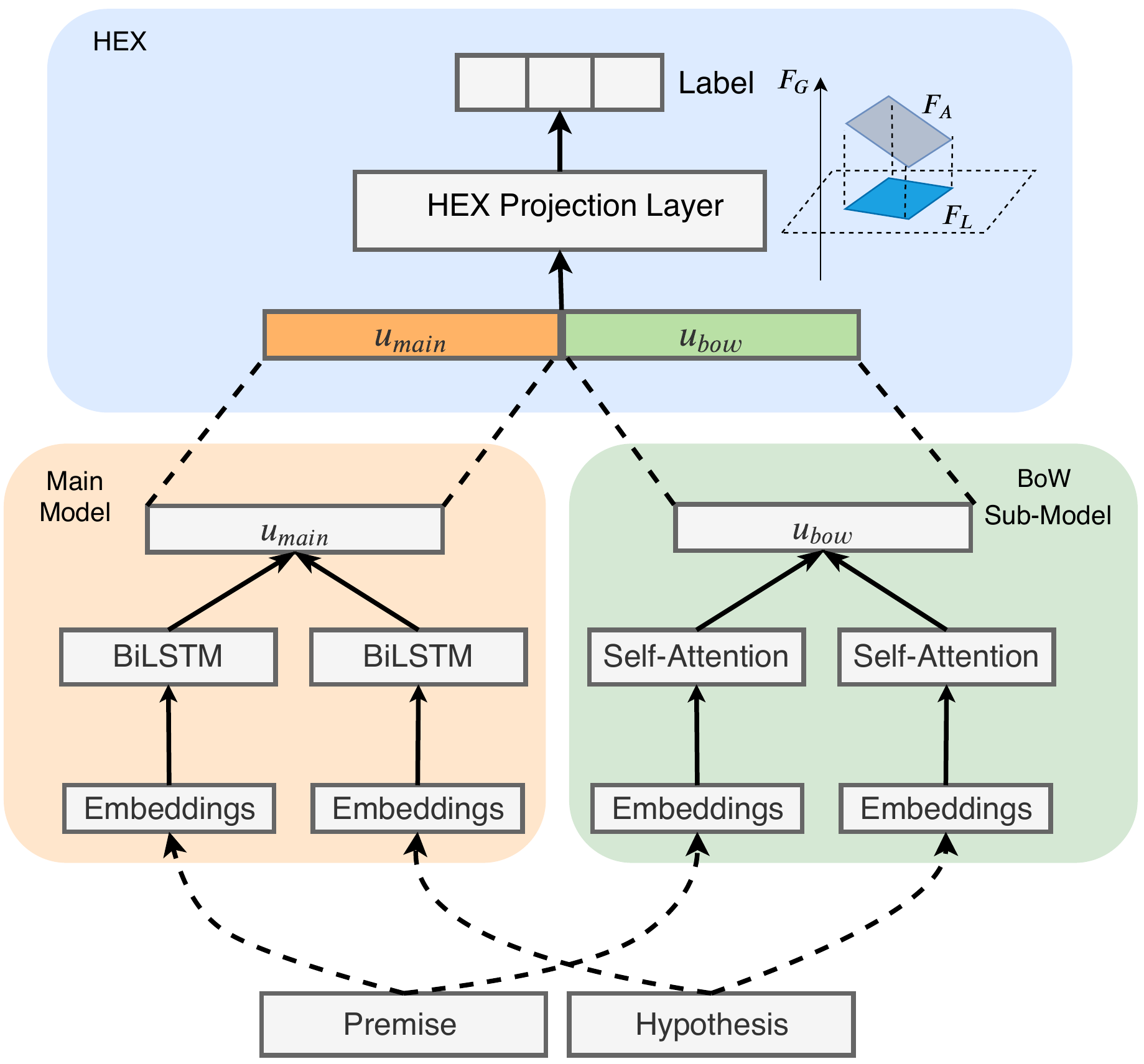}
	\vspace{-8pt}
	\caption{The overall architecture for debiasing the model via orthogonal projection w.r.t. BoW sub-model.}
	\label{fig:hex}
	\vspace{-10pt}
\end{figure}

\vspace{4pt}
\noindent\textbf{BoW Model.} For the BoW sub-model, we first get the embedding of all the words. Then, in order to capture more co-occurrence information of the words, we add a multi-head self-attention layer like the one used in \citet{vaswani2017attention} (but without position embeddings), because we empirically find that this improves the performance. Finally, we use mean-pooling among all the vectors to get the BoW sentence-embedding: $\mathbf{h}_{bow} = \frac{1}{l}\{self\_att(\mathbf{w})\} $.
To get a single representation for the sentence-pair, we used the same concatenation layer as in Eqn~\ref{equ:concat} and pass the vector through an additional MLP to get the representation $u_{bow}$.

\vspace{4pt}
\noindent\textbf{HEX Projection.} Next, in order to encourage the primary model to learn better features that are not learn-able by the BoW model, we used the HEX projection layer from \citet{wang2018learning}, which was originally proposed to improve the domain generalization performance of computer vision models; here we combine HEX with BoW sub-model to robustify NLI models.
With the addition of the BoW sub-model, we can get two representations of the sentence pair $\mathbf{u}_{main}$ and $\mathbf{u}_{bow}$. In order to let the final prediction to use high-level features that are to some extent independent of the shallow and high-biased BoW feature, HEX projection layer projects these two representations into orthogonal spaces to achieve the independence.

The inputs of the HEX projection layers are the BoW model output $\mathbf{u}_{bow}$ and the corresponding output of the main model $\mathbf{u}_{main}$. We use $f$ to denote the final classification network parameterized by $\mathbf{\xi}$. Next, by zero-masking one of the two inputs, the HEX projection layer can receive three different inputs and calculate three different vector outputs:
\begin{equation}
\vspace{-8pt}
\begin{split}
    \mathbf{F}_A=f([\mathbf{u}_{bow};\mathbf{u}_{main}],\mathbf{\xi}) \\
    \mathbf{F}_P=f([\mathbf{0};\mathbf{u}_{main}],\mathbf{\xi}) \\
    \mathbf{F}_G=f([\mathbf{u}_{bow};\mathbf{0}],\mathbf{\xi}) \\
\end{split}
\vspace{-5pt}
\end{equation}
To ensure that the overall model learns different features than the BoW model, we project the joint output $\mathbf{F}_A$ to the orthogonal space of $\mathbf{F}_G$ to get $\mathbf{F}_L$:
\vspace{-5pt}
\begin{equation}
\textbf{F}_L = (\textbf{I}-\textbf{F}_G(\textbf{F}_G^T\textbf{F}_G)^{-1}\textbf{F}_G^T)\textbf{F}_A
\end{equation}
The output learns good representations for both sentences but lies in the orthogonal space of the output got from BoW sub-model's input, thus not over-emphasizing on word-pattern information. This vector goes through the softmax layer to calculate the probabilities for each label. Finally, we follow the original paper \cite{wang2018learning} to minimize a weighted combination of the loss for $\mathbf{F}_L$ and $\mathbf{F}_G$, and use $\mathbf{F}_P$ for testing. 
In Sec. \ref{sec:result_model}, we show that by adding the BoW sub-model orthogonality, the model can be more robust against CWB and WOB while maintaining competitive overall accuracy. Hence, as a response to Q3 in Sec. \ref{sec:intro}, our results indicate that debiasing models at the upper level with regularization on the compositionality is a more promising direction against lexical biases.

\begin{table*}[t!]
\small
\begin{center}
\begin{tabular}{c|ccccc}
\toprule
& \bf MNLI & \multicolumn{2}{c}{\bf Bal} &\multicolumn{2}{c}{\bf Stress*} \\
\bf Train/Test & \bf Acc & \bf Acc  & \bf  Acc\_hr  & \bf Acc &\bf Acc\_hr \\ \midrule
baseline & 69.8 & 70.5 & 45.7 & 50.9 & 38.7\\
+ origin & 69.7/69.2/69.1 & 71.2/71.1/70.6   & 46.3/49.0/47.9 & 49.7/49.2/50.7 & 40.2/40.2/42.1\\
+ synthetic & 69.8/70.0/69.7 & 71.0/70.7/71.2 & 45.7/45.9/47.1 & 67.2/68.5/68.4 & 65.8/68.3/68.4\\
\bottomrule
\end{tabular}
\vspace{-10pt}
\end{center}
\caption{The performance for reducing the CWB via data enhancement/augmentation. The numbers each representing the result after adding 500/20,000/50,000 additional data.}
\label{tab:CWB}
\vspace{-10pt}
\end{table*}

\section{Experimental Setup}
\label{sec:exp_setup}
\subsection{Datasets}
\label{sec:datasets}
We evaluate our models using both off-the-shelf testing datasets as well as new datasets extracted from the original MNLI dataset.
We use the word overlap and the negation sets from the NLI stress tests dataset \cite{naik2018stress}. These two evaluation sets from the NLI stress tests modified the original MNLI development set by appending some meaningless phrases (examples shown in Table~\ref{tab:example}). If the model has certain biases, then the model will be fooled by such perturbations and make the wrong classification. 

In addition, we also extract samples from the original MNLI development dataset to get bias testing sets with exactly the same data distribution. We first select samples that follow the bias pattern from the matched development set. For CWB, we use `not', `no', `any', `never' ,and `anything' as five example contradiction words. To make this testing set balanced for labels (contradiction vs non-contradiction for CWB and entailment vs non-entailment for WOB), we move some samples with the same pattern from the training set to this testing set.\footnote{While this makes our model's performance incomparable to other literature, we train all the models in our experiments in this same setting to ensure the fairness of our analysis comparisons. All our experiments use the same val/test set.} Later we refer to this dataset as \textbf{Bal}.

Since the negation dataset from NLI stress tests dataset only considers the word `not', it fails to evaluate the bias for other contradiction words. We augment this dataset by creating new samples for other contradiction words. We denote the original NLI stress tests dataset as \textbf{Stress} and this augmented one as \textbf{Stress*}. Please refer to the Appendix for a detailed description of how we chose the example contradiction words and created our test sets.

Throughout our experiments, we select the best model during training on the MNLI mismatched development dataset and we tune all the hyper-parameters on the NLI Stress mismatch datasets. All the other datasets are only used as test sets and we only report results on these test sets. We use the MNLI matched development dataset to evaluate the overall performance of the model.

\vspace{-3pt}
\subsection{Metrics}
Overall accuracy is widely used as the only metric for NLI. However, models can get very high accuracy by exploiting the bias patterns. Hence, in order to test how the model performs when it cannot exploit the bias pattern, we focus on model's accuracy on the harder parts of the data (Acc\_hr) where the bias pattern is wrong\footnote{One may wonder if biases can also be evaluated simply using generalization performance. However, good generalization to \emph{current} datasets (e.g., SNLI \cite{bowman2015large}, MNLI \cite{williams2018broad}, SICK \cite{marelli2014semeval}, etc.) is different from being bias-free. As shown in \citet{gururangan2018annotation}, similar annotation artifacts can appear in multiple different datasets. So by overfitting to common lexical biases across multiple datasets, biased models might still reach higher generalization accuracy.}. For the balanced testing set, this subset means samples with  `non-contradiction' label for CWB case and samples with `non-entailment' label for the WOB case. For the NLI stress tests dataset\footnote{Another metric on NLI-stress can be checking the portion of  model predictions on the hard data that is correct both before and after adding the extra words. We empirically verified that this metric shows the same result trends as Acc\_hard.}, this subset means the samples with `non-contradiction' label for the CWB set and the samples with `entailment' label for the WOB set. Ideally, for an unbiased model, it should both have competitive overall performance and perform almost equally well on these harder parts of the data. Hence, we focus on maintaining the accuracy on the whole dataset and improving the Acc\_hr metric. All training details and hyper-parameter settings are presented in Appendix.

\begin{table*}[t!]
\small
\begin{center}
\begin{tabular}{c|c|cccc|cccc}
\toprule
&  & \multicolumn{4}{c|}{\bf CWB} & \multicolumn{4}{c}{\bf WOB} \\
& \bf MNLI & \multicolumn{2}{c}{\bf Bal} &\multicolumn{2}{c|}{\bf Stress*} & \multicolumn{2}{c}{\bf Bal} &\multicolumn{2}{c}{\bf Stress} \\
\bf Model & \bf Acc & \bf Acc  & \bf  Acc\_hr  & \bf Acc &\bf Acc\_hr & \bf Acc  & \bf  Acc\_hr  & \bf Acc &\bf Acc\_hr \\ \midrule
baseline & 70.0 & 70.6 & 45.3  & 49.9 & 37.0   & 75.4 & 58.5  & 59.8 & 40.2 \\
emb\_basic & 67.8 & 70.3 & 49.5  &  50.2 & 41.1 & 73.9 & 56.2  & 56.9 & 35.6 \\
emb\_cond & 67.9 & 68.5 & 46.4  &  48.8 & 38.3 & 74.5 & 54.5  & 56.7 & 39.6\\
sgl\_sent & 67.2 & 68.9 & 47.3  & 48.8 & 37.4 & 73.8 & 55.5  & 54.1 & 29.1 \\
\bottomrule
\end{tabular}
\vspace{-12pt}
\end{center}
\caption{The performance for debiasing the embeddings on CWB and WOB. }
\vspace{-5pt}
\label{tab:embdebias_long}
\end{table*}

\begin{table*}[t!]
\small
\begin{center}
\begin{tabular}{c|c|cccc|cccc}
\toprule
&  & \multicolumn{4}{c|}{\bf CWB} & \multicolumn{4}{c}{\bf WOB} \\
& \bf MNLI & \multicolumn{2}{c}{\bf Bal} &\multicolumn{2}{c|}{\bf Stress*}& \multicolumn{2}{c}{\bf Bal} &\multicolumn{2}{c}{\bf Stress} \\
\bf Model & \bf Acc & \bf Acc  & \bf  Acc\_hr  & \bf Acc &\bf Acc\_hr & \bf Acc  & \bf  Acc\_hr  & \bf Acc &\bf Acc\_hr \\ \midrule
baseline & 69.8$_{\pm\text{0.25}}$ & 70.5$_{\pm\text{0.75}}$ & 45.7$_{\pm\text{2.28}}$ & 50.9$_{\pm\text{1.50}}$  & 38.7$_{\pm\text{3.94}}$ & 76.3$_{\pm\text{0.59}}$ & 59.4$_{\pm\text{0.82}}$ & 58.2$_{\pm\text{3.04}}$ & 37.6$_{\pm\text{9.63}}$   \\
+ BoW & 68.4$_{\pm\text{0.25}}$ & 72.6$_{\pm\text{0.84}}$ & 56.3$_{\pm\text{1.69}}$  & 54.9$_{\pm\text{0.66}}$ & 48.0$_{\pm\text{1.44}}$ & 75.1$_{\pm\text{0.90}}$ & 69.3$_{\pm\text{1.51}}$  & 60.8$_{\pm\text{1.05}}$ & 46.6$_{\pm\text{4.64}}$   \\
\#layers=2 & 69.8$_{\pm\text{0.34}}$ & 69.9$_{\pm\text{0.93}}$ & 44.8$_{\pm\text{1.70}}$ & 51.4$_{\pm\text{0.94}}$  & 40.0$_{\pm\text{2.05}}$ & 76.6$_{\pm\text{0.85}}$ & 58.6$_{\pm\text{0.84}}$ & 58.7$_{\pm\text{1.56}}$ & 40.5$_{\pm\text{5.49}}$   \\
+ BoW & 68.5$_{\pm\text{0.47}}$ & 71.2$_{\pm\text{1.05}}$ & 54.1$_{\pm\text{1.65}}$  & 56.3$_{\pm\text{1.26}}$ & 49.9$_{\pm\text{1.24}}$ & 74.2$_{\pm\text{1.41}}$ & 68.1$_{\pm\text{0.76}}$  & 62.2$_{\pm\text{1.41}}$ & 49.6$_{\pm\text{4.34}}$   \\
\bottomrule
\end{tabular}
\vspace{-10pt}
\end{center}
\caption{The performance for BoW sub-model orthogonality on CWB and WOB. The means and standard deviation here are averaged over five random runs. }
\vspace{-10pt}
\label{tab:hex_long}
\end{table*}

\section{Results}
\label{sec:results}

\subsection{Data-Level Debiasing Results}
\label{sec:result_data}

We first show our baseline's performance on the CWB biases in the first row of Table \ref{tab:CWB}. Since we observe similar performance for CWB and WOB, we leave the results for WOB in Appendix. On every dataset, there's a significant gap between Acc and Acc\_hr, showing the baseline has both strong CWB bias and strong WOB bias.
For the data augmentation/enhancement experiments, we report results after adding 500/20,000/50,000 additional samples. We demonstrate the effect of adding a small portion of data for the 500 case and the limitation of this method using the 20,000 and 50,000 cases.\footnote{Adding additional data (e.g., 50,000) can change the label distribution, but we have experimented with different numbers of additional data between 500 and 50,000 and the reported trend always holds. } 
The results are again shown in Table \ref{tab:CWB}. We use ``+origin" to denote the results from data enhancement using the original dataset and use ``+synthetic" to denote the results from data augmentation by generating new synthetic data similar to NLI stress tests.\footnote{We run all the experiments 5 times and report the mean.}

 With a small number of additional data (500), wherever the data comes from, the performance on the balanced testing set remains very close. However, the performance on the NLI stress tests  improves significantly when it sees 500 synthetic new samples generated in the same way. The gap between the overall accuracy and the Acc\_hr on NLI stress tests is reduced to less than 5\%, which means that the models can easily learn how the synthetic data is generated through only 500 samples. 
Next, we compare the performance after adding 20,000 and 50,000 additional data to check the limitation of the improvement from adding additional data. With this amount of additional original data, the Acc\_hr on the balanced dataset improves and the model is less biased. However, adding 20,000/50,000 synthetic samples doesn't always lead to the improvement on the balanced dataset. This reflects that the generation rules of NLI stress tests dataset are too simple so that training on these adversarial samples is not a good way to robustify the model. However, more natural and diverse synthetic data may be helpful to robustify the models.

There is still a significant gap between overall accuracy and Acc\_hr even after 50,000 samples. Also, the effect of adding the last 30,000 data is very small, indicating a clear limitation of this method. Thus, doing simple data augmentation/enhancement only using the currently available resources is insufficient to fully debias the model. In addition, one has to carefully select which data to add for each different bias, so we need to also design inherently more robust models.

\subsection{Model-Level Debiasing Results}
\label{sec:result_model}

\vspace{4pt}
\noindent\textbf{Debiasing Embeddings (Lower Level Model Debiasing).}  We compared three variants of debiasing embeddings in Table \ref{tab:embdebias_long}. Empirically, we observe that training the whole model with the debias network from a pre-trained baseline can significantly improve the stability of results, so we perform our experiments from one baseline with average performance for fair comparisons. The multi-task coefficient $\lambda$ controls the trade-off between high accuracy and little bias. Here we report the results with $\lambda=1$, which we  find is one good balance point.
From both tables, none of the methods achieved a significant improvement on the Acc\_hr metrics. The best results come from the \textbf{emb\_basic} approach, but even this method only achieves small improvement on the Acc\_hr metric for CWB but does worse on WOB and has a comparable loss on overall Acc. We do not observe any significantly larger improvements with smaller or larger $\lambda$.
We also tried other techniques to further stabilize the training (e.g., freezing the main model when training, using different optimization algorithms), but we observe no significant improvement. 

Therefore, while removing the bias from the embeddings is effective for reducing gender bias (e.g., remove the male bias from the word `doctor' to make the embedding gender-neutral), it does not help in debiasing certain lexical biases. Directly removing information from the embedding only slightly debiases the model but also hurts the overall performance. 
The difference in these results highlights the difference between gender bias and lexical bias problems. As shown in these experiments, lexical biases cannot be effectively reduced at the embedding level. We argue that this is because a majority of lexical biases appear at the compositionality level. For example, for WOB, a biased model will predict ``entailment" entirely relying on the overlapping word embeddings on both sides. Here, even when we make the embeddings completely unbiased, as long as the upper model learns to directly compare the overlapping of embeddings on both sides, there will still exist a strong WOB bias in the model. Hence, in order to robustify models towards lexical bias, we need to develop methods that regularize the upper-interaction part of the model.

\vspace{4pt}
\noindent\textbf{BoW Sub-Model Orthogonality (Higher Level Model Debiasing).} Results for adding the BoW sub-model are shown in Table~\ref{tab:hex_long}. Here, we also show that the improvement trend holds regardless of minor hyper-parameter changes in the model (number of layers). On both CWB and WOB, the model shows a large improvement on Acc\_hr  for both Bal and stress-test datasets.
We achieve close or higher Acc on all the bias testing sets and the overall Acc is only 1.4\%/1.3\% lower than the baseline, showing that adding a BoW sub-model orthogonality will only slightly hurt the model. In conclusion, this approach significantly robustifies the model against CWB and WOB while maintaining competitive overall performance. In comparison to the debiasing embeddings results, we can see that instead of regularizing the content in the word embeddings, regularizing the model's compositionality at the upper interaction level is a more promising direction for debiasing lexical biases. We have also tried combining this method with the data-level debiasing approach above but get no further improvement.\footnote{We also tried some initial simple ensembles of 2 different initializations of BoW sub-models, so that we can potentially regularize against a more diverse set of lexicon biases. When training, the main model is paired with each BoW sub-models to go through each HEX layers and then the output logits are averaged to get the final logits. This ensembling results also outperform the baseline significantly and is higher than the single BoW Sub-Model in WOB Stress, but equal or worse in the other cases. We leave the exploration of different/better ways of ensembling to future work.}

\subsection{Qualitative Feature Analysis}
\begin{figure}[t]
	\includegraphics[width=0.98\linewidth]{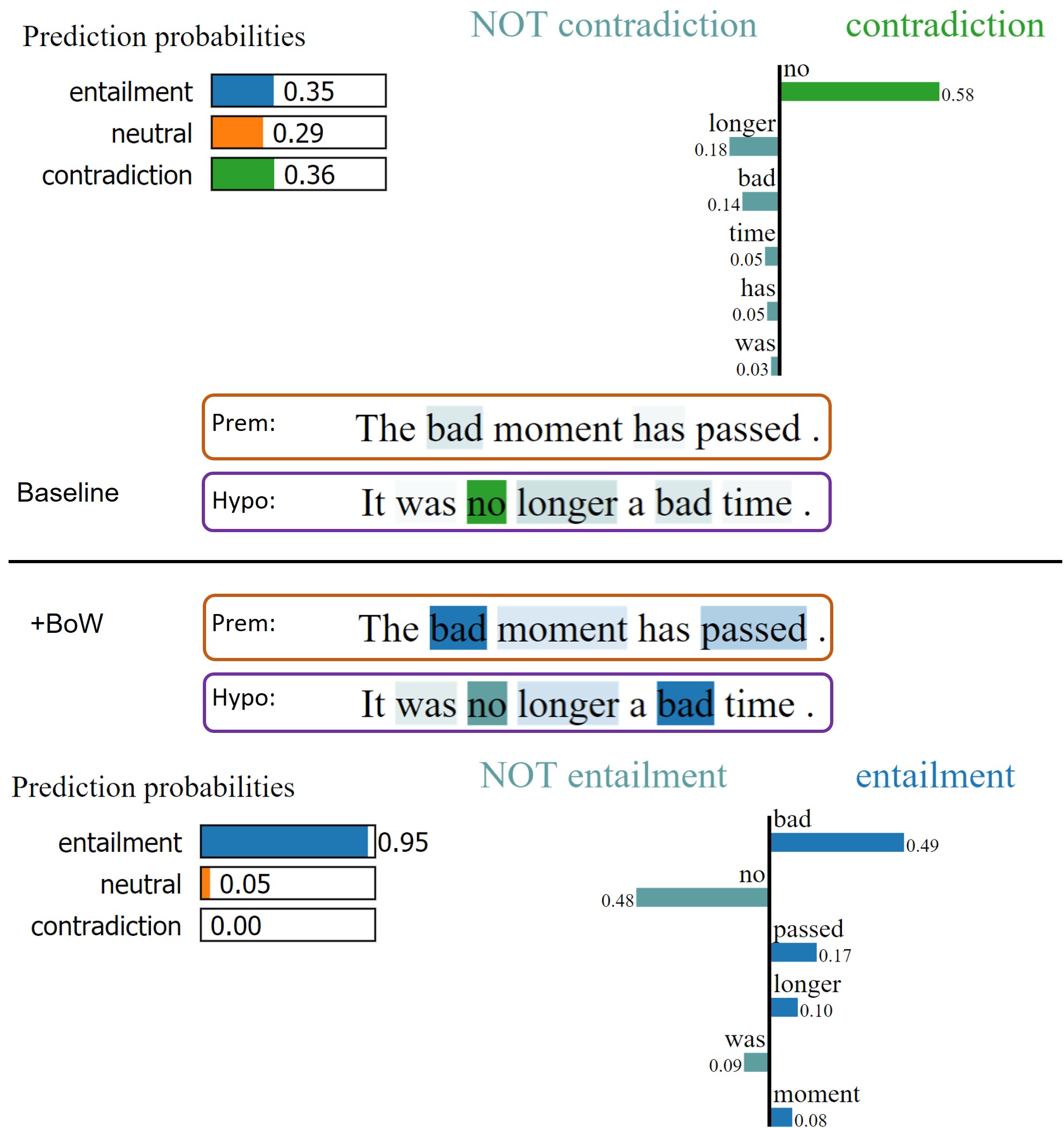}
	\vspace{-8pt}
	\caption{LIME analysis on the CWB Bal dataset showing the 6 most important features used by the model.
	\label{fig:lime_real}}
	\vspace{-10pt}
\end{figure}

We use LIME~\cite{ribeiro2016should} to qualitatively visualize how orthogonal projection w.r.t. BoW sub-model changes the features used by the model. We selected one example from the CWB Bal dataset to see how applying the BoW model with HEX corrects previous mistakes. From Fig. \ref{fig:lime_real}, we can see that before applying the BoW sub-model (the upper part of the figure), the model predicts the contradiction label almost solely based on the existence of the word ``no" in the hypothesis. However, after applying our BoW sub-model with HEX projection, our model can give higher importance to other useful features (e.g., the match of the two ``bad" tokens, and the match of important past-tense temporal words such as ``passed'' and ``longer'' in the premise-hypothesis pair) despite the fact that ``no" still has high influence towards the contradiction label. Another example from the CWB Stress* dataset can be seen in Appendix.

\section{Conclusion}
\label{sec:conclusion}

We study the problem of lexical dataset biases using WOB and CWB as two examples. We first showed that lexical dataset biases cannot be solved by simple dataset changes and motivate the importance of directly designing model-level changes to solve this problem. For model-level changes, we first show the ineffectiveness of embedding-debiasing approaches, thus highlighting the uniqueness of lexical bias against gender bias problems. Next, we robustify the model by forcing orthogonality between a BoW sub-model and the main model and demonstrate its effectiveness through several experiments. Since none of our methods is bias-type specific, we believe these results can also be generalized to other similar lexical biases. 
Finally, we would like to point out that our methods and results here do not mean to belittle the importance of collecting clean/unbiased data. 
We strongly believe in the importance of unbiased data for model design and evaluation. However, some biases are inherent and inevitable in the natural distribution of the task (e.g., for NLI, it is natural that sentences with high overlapping are most likely entailment pairs). Therefore, our work stresses that it is also very important to encourage the development of models that are unlikely to exploit these inevitable biases/shortcuts in the dataset. Neither model-level debiasing nor data-level debiasing alone is the conclusive solution for this problem. Joint efforts are needed for promoting unbiased models that learn true semantics; and we hope our paper can encourage more work towards this important direction.

\vspace{-3pt}
\section*{Acknowledgments}
\vspace{-3pt}
We thank Snigdha Chaturvedi, Shashank Srivastava, and the reviewers for their helpful comments. This  work was
supported by DARPA YFA17-D17AP00022, NSF-CAREER Award 1846185, ONR Grant N00014-18-1-2871. The views in this article are the authors', not of the funding agency.

\bibliography{emnlp-ijcnlp-2019}
\bibliographystyle{acl_natbib}

\section*{Appendix}
\appendix

\section{Training Details}
For all our models except BERT \cite{devlin2018bert}, we use pre-trained 300-dimension GloVe \cite{pennington2014glove} word embeddings to initialize the embedding layers. The hidden dimension of LSTM \cite{hochreiter1997long} is 300. We use Adam \cite{kingma2014adam} as the optimizer and the initial learning rate is set to 0.0004. We apply dropout \cite{srivastava2014dropout} with a rate of 0.4 to regularize our model. For the model with HEX projection, we apply all the tricks in the original paper \cite{wang2018learning} (column-wise normalize the input features in every batch, fine-tune from a trained model with the bottom layer fixed) to stabilize the training. In our experiments, we set the multi-task coefficient between loss for $F_L$ and $F_G$ to 1.0 and 0.3.

\section{Detailed Description of the Extraction of Balanced Testing Sets}
\subsection{Extraction of the Contradiction-Word-Bias Testing Set}
For evaluating the contradiction-word-bias (CWB), we look for words that both have a strong bias towards the `contradiction' label and have a significant number of samples in the training set. We first select `no', `any', `never' and `anything', which are four most frequent words with over 50\% of samples in the training data containing these words labeled as 'contradiction'. Since most of the analysis papers also study the bias of `not', here we also include the `not' as the contradiction word. However, as in the training set of MNLI \cite{williams2018broad}, only 45.3\% of the samples are `contradiction', so the bias of `not' is actually not as strong as the other words.

Next, in order to create a balanced dataset for these selected contradiction-words, we first select the samples containing these words from the matched development set. In order to let the samples be more difficult and better test the model's bias. We only select the samples where the hypothesis samples contain the contradiction word, while there's no negation word in the premise sentence (so that the contradiction word is generated by the annotator instead of copying from the premise sentence). Since the bias of `not' is not uniformly strong, here we only select samples that both contain `not' and have small Jaccard distance \cite{hamers1989similarity} between the sentence pairs, which we empirically find that the bias is stronger.

After selecting these samples, we can extract a testing set with most of the samples labeled as contradiction, but the label distribution is severely unbalanced. In order to balance the label distribution, we randomly sample some examples from the training set using the same criterion (containing contradiction word in the hypothesis sentence but no negation word in the premise sentence) and put them in the testing set.
Our resulting dataset contains 1100 samples with 550 are labeled as contradiction and the other 550 are non-contradiction labels.
Since the domain of the training set is different from the domain of the mismatched validation set, we only extract a balanced test set based on the matched validation set.

\begin{table}[t]
\small
\centering
\begin{tabular}{|c|c|}
\hline
\bf contradiction-word & \bf appended phrase \\ \hline
no & and false is no true \\ \hline
any & and any true is true \\ \hline
never & and false is never true \\ \hline
anything & and anything true is true \\ \hline
not & and false is not true \\ \hline
\end{tabular}
\caption{The phrases to append at the end of the hypothesis sentence for each contradiction word.}
\label{tab:app_syn}
\end{table}

\begin{table*}[t!]
\small
\begin{center}
\begin{tabular}{c|ccccc}
\toprule
& \bf MNLI & \multicolumn{2}{c}{\bf Bal} &\multicolumn{2}{c}{\bf Stress} \\
\bf Train/Test & \bf Acc & \bf Acc  & \bf  Acc\_hr  & \bf Acc &\bf Acc\_hr \\ \midrule
baseline & 69.8 & 76.3 & 59.4 & 58.2 & 37.6\\
+ origin &70.1/70.0/69.4  & 77.1/77.5/76.4 & 61.5/64.1/64.7 & 56.0/58.0/55.4 & 31.0/37.3/29.5 \\
+ synthetic & 70.0/69.8/69.6 & 77.2/75.7/75.7 & 61.3/58.8/58.6 & 67.7/68.8/68.7 & 66.2/72.9/72.0\\
\bottomrule
\end{tabular}
\vspace{-10pt}
\end{center}
\caption{The performance of LSTM baseline model for reducing the WOB via data enhancement/augmentation. The numbers each representing the result after adding 500/20,000/50,000 additional data.}
\vspace{-5pt}
\label{tab:WOB}
\end{table*}

\begin{table*}[ht]
\small
\begin{center}
\begin{tabular}{c|ccccc}
\toprule
& \bf MNLI & \multicolumn{2}{c}{\bf Bal} &\multicolumn{2}{c}{\bf Stress*} \\
\bf Train/Test & \bf Acc & \bf Acc  & \bf  Acc\_hr  & \bf Acc &\bf Acc\_hr \\ \midrule
baseline & 82.3 & 84.2 & 71.2 & 55.8 & 41.9\\
+ origin & 82.3/82.6/82.7 & 83.8/83.7/83.6   & 70.7/70.6/70.2 & 55.7/55.3/55.2 & 42.4/41.7/43.2\\
+ synthetic & 82.6/82.4/82.4 & 84.3/84.1/84.3 & 71.9/71.2/71.5 & 83.3/84.0/83.9 & 81.9/83.2/83.0\\
\bottomrule
\end{tabular}
\end{center}
\vspace{-10pt}
\caption{The performance of BERT for reducing the CWB via data enhancement/augmentation. The numbers each representing the result after adding 500/20,000/50,000 additional data.}
\label{tab:CWB_BERT}
\vspace{-5pt}
\end{table*}

\begin{table*}[t!]
\small
\begin{center}
\begin{tabular}{c|ccccc}
\toprule
& \bf MNLI & \multicolumn{2}{c}{\bf Bal} &\multicolumn{2}{c}{\bf Stress} \\
\bf Train/Test & \bf Acc & \bf Acc  & \bf  Acc\_hr  & \bf Acc &\bf Acc\_hr \\ \midrule
baseline & 82.3 & 90.5 & 87.0 & 58.1 & 6.49\\
+ origin &82.7/82.4/82.4  & 91.3/90.5/90.8 & 87.9/87.2/87.5 & 58.1/58.2/58.1 & 7.43/7.61/5.88 \\
+ synthetic & 82.4/82.5/82.5 & 90.7/90.6/91.1 & 87.0/86.7/87.5 & 83.4/84.0/83.9 & 82.4/83.8/83.8\\
\bottomrule
\end{tabular}
\end{center}
\vspace{-10pt}
\caption{The performance of BERT for reducing the WOB via data enhancement/augmentation. The numbers each representing the result after adding 500/20,000/50,000 additional data. }
\label{tab:WOB_BERT}
\vspace{-5pt}
\end{table*}

\subsection{Extraction of the Word-Overlapping-Bias Testing set}
We first sort the samples in the MNLI matched validation set using Jaccard distance \cite{hamers1989similarity} and choose the samples with the smallest distance (highest overlapping). In order to match the size of the contradiction-word-bias testing set, we select the top 550 samples with entailment label and the top 550 samples with non-entailment label to get a dataset with high word overlapping but balanced label distribution.

\section{Construction of Synthetic Data}
We follow the construction rule of the NLI stress tests \cite{naik2018stress} to generate synthetic data for the training set. We appended meaningless sentences at the end of the hypothesis sentence and keep the original label unchanged.
For CWB, we focus on 5 different contradiction words: `no', `any', `never', `anything' ,and `not'. Therefore, for each sentence pair, we create five different new pairs by appending five different phrases for evaluating the bias of each contradiction word. The appended phrases are listed in Table \ref{tab:app_syn}.
For WOB, we also follow \cite{naik2018stress} to append `and true is true' to every hypothesis sentence to create one new pair for each sample.

\begin{figure*}[t!]
\centering
	\includegraphics[width=0.95\linewidth]{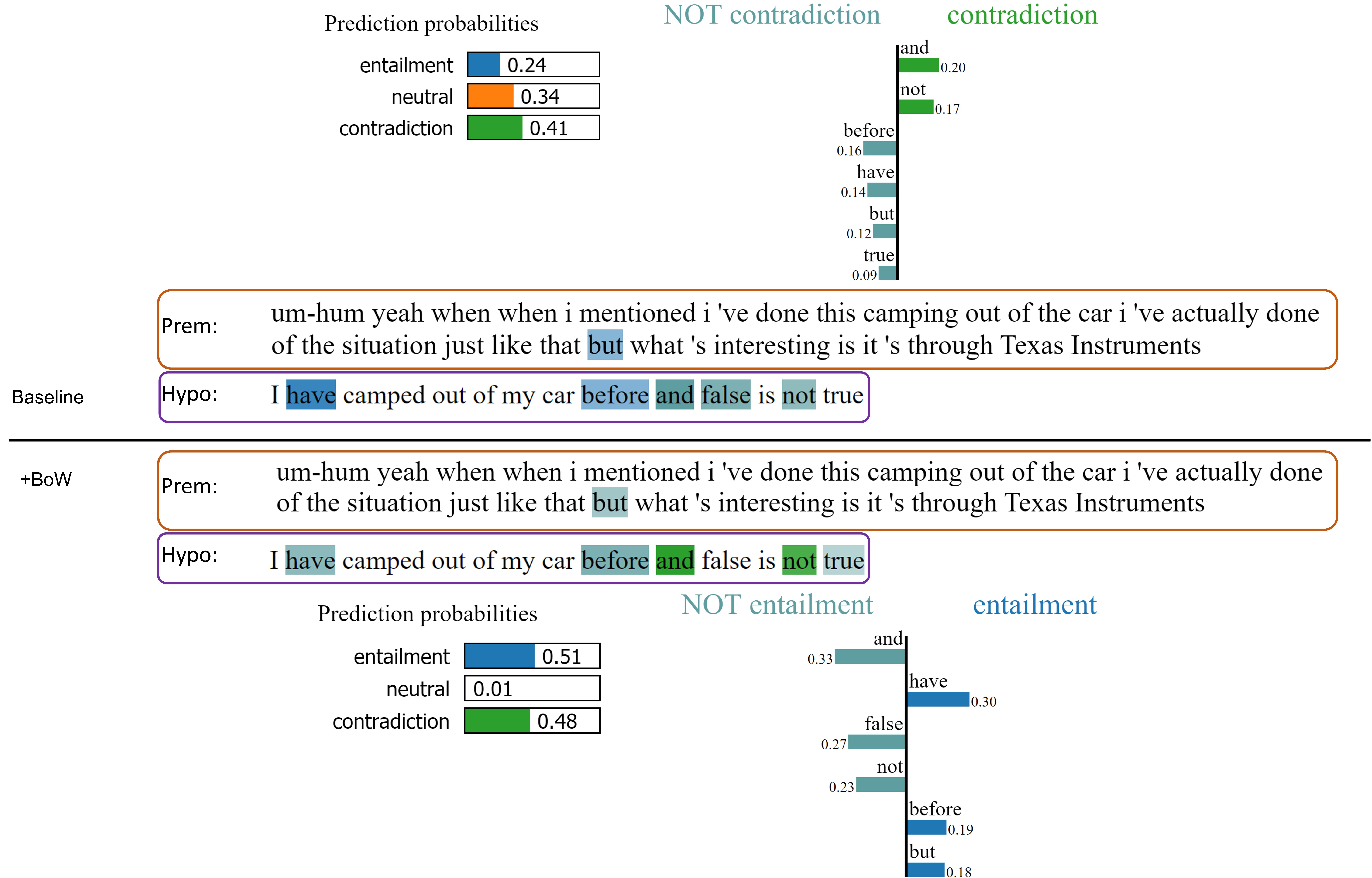}
	\vspace{-8pt}
	\caption{LIME analysis on the CWB Stress* dataset showing the 6 most important features used by the model.}
	\vspace{-7pt}
	\label{fig:lime_synt}
\end{figure*}

\section{Data Augmentation/Enhancement Results for BERT}
The data augmentation/enhancement results for BERT-base \cite{devlin2018bert} is shown in Table \ref{tab:CWB_BERT} and Table \ref{tab:WOB_BERT}. \footnote{We run all the experiments 5 times and report the mean.}
As is shown in Table \ref{tab:CWB_BERT}, BERT shows significant performance gap between Acc and Acc\_hr on both CWB datasets, indicating BERT's clear bias on CWB. As for WOB, the gap between Acc and Acc\_hr for Bal is much smaller, however, the performance on Stress is very poor. Therefore, we assume that even though BERT achieves a high score on the WOB Bal dataset, BERT is just overfitting the dataset in another different way, i.e., there is still significant WOB bias in BERT. In conclusion, in our experiment, BERT still shows significant CWB and WOB.

Similar to our main data augmentation/enhancement results, here we find that after adding 500 additional synthetic samples, BERT can quickly learn their pattern. But still, adding more synthetic data doesn't help improve the performance on the Bal dataset. For BERT, we also cannot see any significant improvement when adding additional original samples. In all the + origin experiments, BERT performs similarly. Again, this shows the limitation of the data augmentation/enhancement approach, especially starting with a stronger baseline as BERT.

\vspace{-3pt}
\section{More Qualitative Feature Analysis}
\vspace{-3pt}
In Fig. \ref{fig:lime_synt}, we can see the feature importance change before/after adding the BoW sub-model for a CWB Stress* example (we chose a borderline example where the prediction distribution change to the correct label is not extreme). We can see that before adding the BoW sub-model orthogonality-projection, the extra misleading words (both ``and" and ``not") confused the model to predict the wrong contradiction label, while after adding the BoW sub-model, our model can assign higher weights to useful features such as ``have", ``before", etc.

\end{document}